\pdfoutput=1

\documentclass[11pt]{article}

\usepackage[preprint]{acl}

\usepackage{times}
\usepackage{latexsym}
\usepackage{amssymb}

\usepackage[T1]{fontenc}

\usepackage[utf8]{inputenc}

\usepackage{microtype}

\usepackage{inconsolata}

\usepackage{graphicx}

\usepackage{hyperref}
\usepackage{url}
\usepackage{xspace}
\usepackage{xcolor}
\usepackage{wrapfig}
\usepackage{booktabs}
\usepackage[normalem]{ulem}
\usepackage{arydshln}
\usepackage{caption}
\usepackage{multirow}
\usepackage{graphicx}
\usepackage{hyperref}
\usepackage{textcase}
\usepackage{fancyvrb}
\useunder{\uline}{\ul}{}
\definecolor{SkyBlue}{rgb}{0.53, 0.81, 0.92}
\definecolor{LightGreen}{RGB}{144, 238, 144}
\definecolor{Pottery}{RGB}{221, 126, 107}

\title{Contextual Experience Replay for Self-Improvement of Language Agents}

\author{%
Yitao Liu $^{\heartsuit\spadesuit}$
\thanks{This work was done while Yitao was a student researcher at Princeton University} \quad
Chenglei Si $^\diamondsuit$ \quad 
Karthik Narasimhan $^\heartsuit$ \quad
Shunyu Yao $^\heartsuit$ \quad \\
$^\heartsuit$Princeton University
\quad
$^\spadesuit$The University of Hong Kong
\quad
$^\diamondsuit$Stanford University \\
  \texttt{lyitao17@gmail.com} \\
}

\newcommand{\ourmethod}{\textsc{CER}\xspace}

\begin{document}
\maketitle
\begin{abstract}
Large language model (LLM) agents have been applied to sequential decision-making tasks such as web navigation, but without any environment-specific experiences, they often fail in these complex tasks. 
Moreover, current LLM agents are not designed to continually learn from past experiences during inference time, which could be crucial for them to gain these environment-specific experiences.  
To address this, we propose Contextual Experience Replay (\ourmethod), a training-free framework to enable efficient self-improvement for language agents in their context window. Specifically, \ourmethod accumulates and synthesizes past experiences into a dynamic memory buffer. These experiences encompass environment dynamics and common decision-making patterns, allowing the agents to retrieve and augment themselves with relevant knowledge in new tasks, enhancing their adaptability in complex environments. We evaluate \ourmethod on the challenging \textsc{WebArena} and \textsc{VisualWebArena} benchmarks. 
On \textsc{VisualWebArena}, \ourmethod achieves competitive performance of 31.9\%.
On \textsc{WebArena}, \ourmethod also gets a competitive average success rate of 36.7\%, relatively improving the success rate of the GPT-4o agent baseline by 51.0\%.
We also conduct a comprehensive analysis on it to prove its efficiency, validity and understand it better.
\end{abstract}

\section{Introduction}



Building an autonomous agent that can help with people's daily tasks has been a long-standing goal of artificial intelligence research \citep{Russell1995AI, Franklin1996AgentTaxonomy}. Recently, large language models \citep{Achiam2023GPT4, Anthropic2024Claude3, Gemini2023} have shown impressive performance in text \citep{hendrycks2021measuring} and code generation \citep{chen2021evaluatinglargelanguagemodels,xie2024textreward}, reasoning \citep{wei2022chain, Yao2023TreeOT}, and decision-making tasks \citep{yao2022react, zhou2024language, xu2024lemurharmonizingnaturallanguage, xie2024openagents}, which paves the way for building an agent to automate computer tasks. 
On two realistic web navigation benchmarks, \textsc{WebArena} \citep{zhou2024webarenarealisticwebenvironment} and \textsc{VisualWebArena} \citep{koh2024visualwebarenaevaluatingmultimodalagents}, humans can achieve success rates of 78.24\% and 88.70\%, correspondingly. However, the current methods, with the most frontier models, can only achieve a success rate of around or less 20\% without human involvement. 

\begin{figure*}[t]
    \centering
    \includegraphics[width=1.0\linewidth]{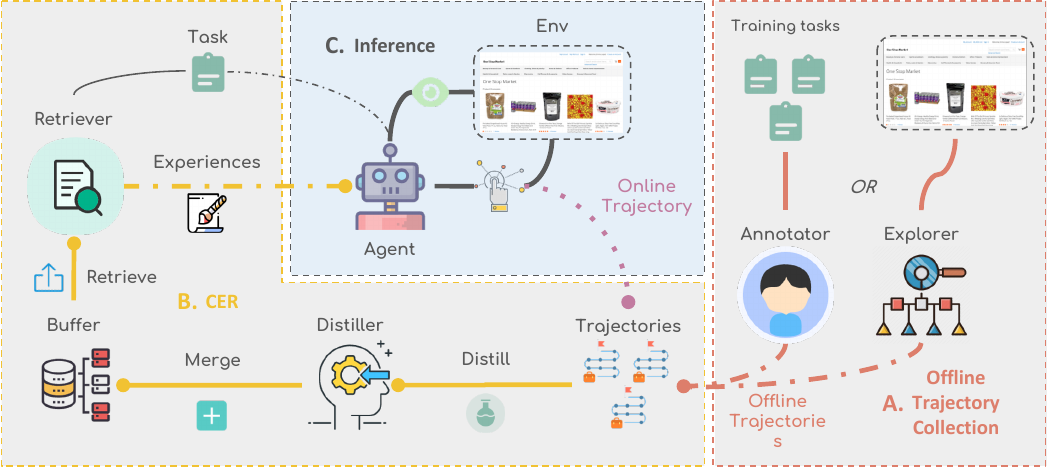}
    \caption{Overview of Contextual Experience Replay including offline and online settings. 
    (1) In the online setting, it will start from stage {\color{gray}{C}} and loop between stage {\color{gray}{C}} and {\color{yellow}{B}} for each task, i.e. solve task $i$, learn experiences from it and solve task $i+1$ with previous experiences, and so on. (2) In the offline setting, stage {\color{Pottery}{A}} is needed to get offline trajectories, then it goes from stage {\color{yellow}{B}} to {\color{gray}{C}} and finally stays in stage {\color{gray}{C}}, i.e., learns experiences from offline trajectories and solves all tasks. (3) In the hybrid setting, it will begin from stage {\color{Pottery}{A}} and loop between {\color{yellow}{B}} and {\color{gray}{C}}, conducting both offline and online learning.
    }
    \label{fig:off_on}
\end{figure*}

One important reason is the lack of prior knowledge of each environment, which is critical for such difficult multi-step task solving in the complex web environment. While training in each specific environment is costly, current language agents seldom have an efficient way to continually learn about the environment, so they need to explore the environment from scratch for every single task \citep{koh2024treesearchlanguagemodel}.

In this work, we propose Contextual Experience Replay (\ourmethod), a simple and effective framework to enable the self-improvement of language agents in complex environments. 
\ourmethod is loosely inspired by experience replay \citep{schaul2016prioritizedexperiencereplay, rolnick2019experiencereplaycontinuallearning}, an important algorithm in reinforcement learning which highlights storing past trajectories into a buffer and training the agent with these data. 

\begin{figure*}[t]
    \centering
    \includegraphics[width=1.0\linewidth]{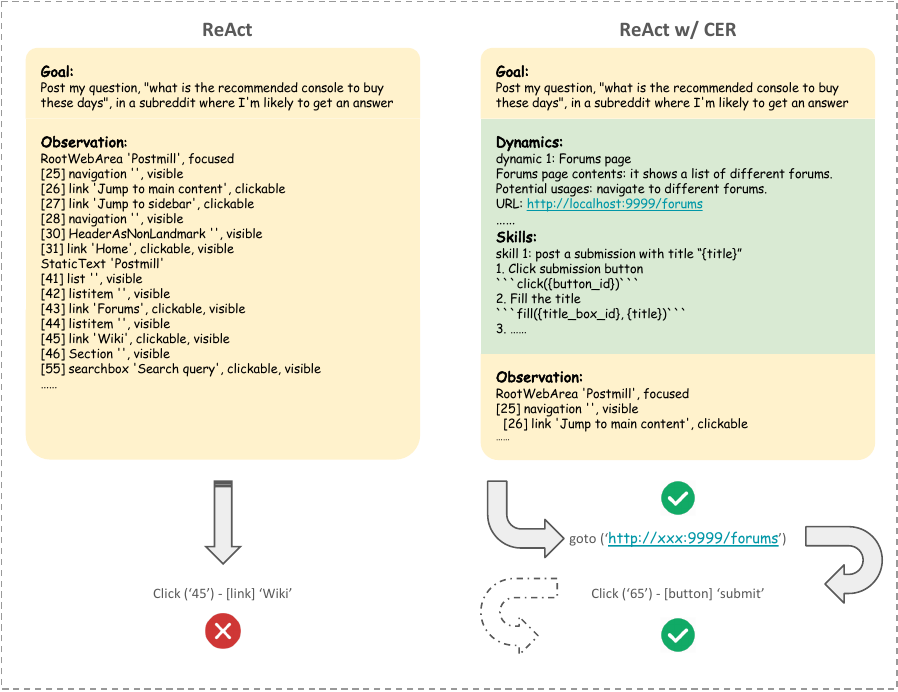}
    \caption{Compare ReAct baseline with ReAct + \ourmethod. The {\color{LightGreen}{experiences}}, including dynamics and skills, are obtained through multiple modules as in Fig.\ref{fig:off_on}. They are "replayed" in the context window of the model, helping the agent to make correct decisions. For simplicity, the thinking process is neglected in the figure.}
    \label{fig:example}
\end{figure*}

Our approach allows agents to distill experience from trajectories, including environment dynamics and common decision-making patterns, from past trajectories, store them into a dynamic memory, retrieve them with the current task, and replay them in context when solving new tasks. Fig.\ref{fig:off_on} shows how \ourmethod works under different settings. Online, offline, and hybrid settings are divided by the source of trajectories, i.e., the time to get the trajectories. As in Fig.\ref{fig:off_on}, in the online setting, the agent will start from the inference stage (C) without any experience. After completing a task, \ourmethod gets the (online) trajectory from it, distills experiences from the trajectories, and merges it into the buffer. During the inference of the next task, the agent will be augmented with retrieved helpful experiences and so on. In the offline setting, a set of trajectories will be collected in advance (stage A), distilled into experiences, and stored. Then, the agent will solve tasks on the test set with retrieved experience from the fixed buffer. The hybrid setting is the combination of these two, i.e., going through the offline learning stage before online learning. Fig. \ref{fig:example} also shows how the experience is utilized by the agent with an example.


We evaluated \ourmethod on two realistic web benchmarks \textsc{WebArena} \citep{zhou2024webarenarealisticwebenvironment} and \textsc{VisualWebArena} \citep{koh2024visualwebarenaevaluatingmultimodalagents}. \ourmethod improves the GPT-4o baseline by a large margin and achieves competitive results on these two benchmarks while orthogonal with most other methods. On \textsc{WebArena}, \ourmethod shows a relative improvement of 51.0\% over the GPT-4o baseline and achieves an overall success rate of 36.7\%, competitive with other state-of-the-art (SOTA) methods. On \textsc{VisualWebArena}, \ourmethod outperforms the tree search-based method by 20.8\% in relative performance with dozens of times fewer token costs and achieves a competitive success rate of 31.9\%. 




We conducted further analysis to investigate the improvements of \ourmethod with various metrics, such as cross-template success rate, stability (preservation of old
knowledge), and plasticity (acquisition of new knowledge) \citep{Grossberg1982CognitiveCode, rolnick2019experiencereplaycontinuallearning} (\S\ref{subsec:improvement_analysis}, \S\ref{subsec:stability}), demonstrating its generalizability and effectiveness as a self-improvement system. Also, we show that through the combination with a sampling-based method, \ourmethod pushes the boundary forward again, showing its compatibility with other methods (\S\ref{subsec:synergy}).

In summary, our contributions are as follows:
\begin{itemize}
    \item We propose a simple but effective self-improvement framework for language agents: \ourmethod. {\ourmethod distills fine-grained skills and environment dynamics from both successful and failed trajectories. Importantly, it works for offline, online, and hybrid settings.}
    \item CER shows competitive performance on multimodal web navigation tasks. It also shows excellent stability and plasticity \citep{Grossberg1982CognitiveCode, rolnick2019experiencereplaycontinuallearning}, as well as good synergy with other off-the-shelf methods.
    \item We do a comprehensive analysis on \ourmethod to prove its validity and understand the improvements better. 
\end{itemize}

\section{Related work}
\paragraph{LLM Agents}
The increasing capabilities of LLMs have enabled new applications where agents built with LLMs can take action and interact with external environments. 
To enable action-taking, methods like ReAct~\citep{yao2022react} prompt LLMs to interleave actions and reasoning in the output. Apart from action-taking, planning and search is also an important component for agents. Methods like Reflexion~\citep{Shinn2023ReflexionLA}, Self-Refine~\citep{Madaan2023SelfRefineIR}, Tree-of-Thought~\citep{Yao2023TreeOT}, and Tree Search~\citep{koh2024treesearchlanguagemodel} \citep{zhou2024language} enable LLMs to revise their reasoning and perform deliberate search among their action space. 

\paragraph{Web Agent Environments}
LLM agents are increasingly being employed to perform various digital tasks on behalf of humans, with interacting with websites being a common application area supported by numerous benchmarks. For instance, WebShop~\citep{yao2023webshopscalablerealworldweb} tasks agents with identifying products that meet specific user requirements by interacting with e-commerce platforms. Extensions such as WebArena~\citep{zhou2024webarenarealisticwebenvironment} and Mind2Web~\citep{deng2023mind2webgeneralistagentweb} have broadened the scope of tasks to include a wider variety of websites and more realistic applications, encompassing activities like trip booking, information retrieval, website navigation, and social media management. VisualWebArena~\citep{koh2024visualwebarenaevaluatingmultimodalagents} designs challenging multimodal web navigation tasks that require agents to leverage visual grounding and understand image inputs. Among these benchmarks, WebArena and VisualWebArena provide the most realistic, controllable, and interactable environments, which makes the tasks more challenging and the results reproducible. The interactive characteristics are also beneficial for our self-improvement paradigm. 

\paragraph{Learning from Memory or Past Experiences}
Some previous works have investigated the storage of memories of past agent trajectories. Generative agents~\citep{Park2023GenerativeAI} use similar strategies to investigate human behaviors with such a human-like strategy. Voyager \citep{wang2024voyager} enables the agent to learn diverse skills in Minecraft.

Similarly, frameworks such as ExpeL~\citep{Zhao2023ExpeLLA} and Synapse~\citep{Zheng2023SynapseTP} leverage stored past task trajectories as memory, which are dynamically retrieved to support task execution. However, they either test on relatively simple web environments \citep{yao2023webshopscalablerealworldweb} or use raw and long observation-action pairs as exemplars directly, which limits their applicability to more complex environments. Learn-by-interact~\citep{su2025learnbyinteractdatacentricframeworkselfadaptive} uses related documentation to generate tasks and learn through bridging the misalignment between task instructions and generated trajectories.
Besides, AutoGuide~\citep{fu2024autoguide}, AutoManual~\citep{chen2024automanual}, and Agent Workflow Memory~\citep{Wang2024AgentWM} apply similar ideas. However, their mechanism requires them to obtain the ground truth reward before they can function, which significantly limits their applicability. Also, their mechanisms are relatively simple and not designed for self-improvement in real-world scenarios. For example, Agent Workflow Memory does not have a retrieval module and updates the whole workflow memory in a rewriting style. 
In our work, we construct a well-designed, efficient, and scalable self-improvement framework for autonomous language agents and test it in two challenging and realistic web environments. The experiences contain both environment dynamics and decision-making patterns. We also investigate it both qualitatively and quantitatively and demonstrate its advantage in terms of applicability in different learning paradigms and compatibility with other agent methods. 

\section{\ourmethod: Contextual Experience Replay}
Consider a general setup for a language agent $A$, powered by a language model $M$ with a context window $C$, to solve a sequential decision-making task in an environment. 
\ourmethod includes four separate modules: distillation module $D$, retrieval module $R$, dynamic experience buffer $\epsilon$ and the base decision-making agent itself $A$ as shows in Fig.\ref{fig:off_on}. \ourmethod can start working given a arbitrary set of trajectories $\mathbb{T} = \{\tau_1, \tau_2, \dots, \tau_n\}$. All modules here are implemented by prompting a visual language model (VLM), i.e. GPT-4o in our implementation. Details of prompts for each module can be found in \ref{subsec:cer_prompts}. 
\subsection{Distill experiences from trajectories}
Given a trajectory set $\mathbb{T}$, the distillation module will distill experiences $\mathbb{E} = \{E_1, E_2, \dots, E_n\}$ from them one by one where $E_i = (D_i, S_i)$. $D_i$ stands for environment dynamics, or dynamics in short, and $S_i$ represents useful decision-making patterns, or skills in short. The dynamics provide useful state information to help the agent make state-aware decisions or directly navigate to the state through its URL. These skills provide common decision-making patterns, inspiring agents to take better action. Fig. \ref{fig:example} provides an example for them.
We use two separate modules for the distillation of dynamics and skills due to their different characteristics. The detailed prompts are available in \ref{subsec:cer_prompts}.
The output format is similar to ReAct \citep{yao2022react}, asking the model to issue a think action before outputting each distillation. The dynamics distillation module will distill a list of summaries of different web pages, their corresponding URL, and inferred possible usages. The skill distillation module is instructed to summarize a list of useful skills. Each of them includes a brief overall summary (e.g. Navigate to forum \{\texttt{forum name}\}) and the corresponding detailed step-by-step guidelines. Specifically, the guideline contains both natural language summaries and concrete action examples for each step, as the example in Fig. \ref{fig:example}. While the natural language summaries provide flexible and general high-level instruction, the example helps the agent to understand the step and also format its output better. The model is required to output the final distillation in an abstract and general way, i.e. navigate to forum \{\texttt{forum name}\} instead of navigate to forum "books", to ensure that the experiences can be broadly applied. The model is also provided with existing experiences in the buffer to avoid repetitive distillation, allowing the continual accumulation of the experiences across time. 
\subsection{Retrieve experiences from buffer}
After the distillation period, the buffer $\epsilon$ now includes a set of useful experiences $\mathbb{E} = \{E_1, E_2, \dots, E_n\}$. Similar to the distillation module, we designed two separate modules to retrieve dynamics and skills correspondingly. Each module is implemented by prompting a VLM. We prompt the model with general instructions, the current task goal, the website descriptions, and all dynamics or skills available in the buffer. Then, the model will output the top-$k$ useful and informative experiences and pass this to the language agent. The prompts can be found in \ref{subsec:cer_prompts}. This module makes it possible for the distillation module to continuously merge new experiences and help the agent filter out useful experiences for the current task.

\subsection{Decision-making with Contextual experience replay}
To best utilize the in-context learning capability of language models, we transform the selected $k$ experiences $\mathbb{E} = \{E_1, E_2, \dots, E_k\}$ into natural language experience descriptions $E_{NL} = f(\mathbb{E})$ through a programmatic mapping f and integrate them into the model's context $C$, resulting in a new augmented context $C' = g(C, E_{NL})$. Therefore, the decision-making policy underneath will be influenced by the additional experiences, and the agent $A$ can issue better actions with reference to the experiences. The context comparison between the baseline agent and \ourmethod is shown in Fig.\ref{fig:example}.
\subsection{Combination of offline and online learning}
\label{subsec:off_on_method}
The source of the trajectories to learn from is important for \ourmethod. Depending on the source of trajectory data, \ourmethod can be divided into offline, online, and hybrid versions. Online data is collected from past task-solving trajectories in the environment during inference time. Specifically, in the online setting, there are no trajectories provided at the very beginning, but as the procedure goes on, there will be self-generated trajectories from past tasks. \ourmethod will run the distillation module after each task and run the retrieval and replay module in the next task. Different from an online setting, offline learning means there is a training set of trajectories at the beginning for \ourmethod to learn from but no further learning during inference. Additionally, these two settings can be combined to serve as a whole system, i.e., learn from a fixed training set first and then self-evolve in the environment with self-generated data.

\section{Experiments}

We evaluate \ourmethod on the full set of \textsc{WebArena} \citep{zhou2024webarenarealisticwebenvironment} (WA) in offline, online, and hybrid settings. For \textsc{VisualWebArena} \citep{koh2024visualwebarenaevaluatingmultimodalagents} (VWA), we only evaluate in an online setting for cost consideration. The reason we chose these two is that they provide interactive, realistic, and reproducible web environments that are better for applying self-improvement and still close to real-world scenarios. \textsc{WebArena} have 812 tasks across five different websites corresponding to different domains: shopping, shopping administration, online forum, map, and project collaboration (Gitlab). \textsc{VisualWebArena} retains the shopping and forum website, adds another classifieds website, and designs 910 tasks on top of them. Although they share two websites, the focus of their tasks is different. Most of the tasks in \textsc{WebArena} only have text descriptions of task goals, while a large portion of tasks in \textsc{VisualWebArena} involve visual input as part of task goals and require an understanding of the visual information of the current website. This also leads to their large variations of task types.

\subsection{Implementation Details}

\begin{table*}[t]
\centering
\renewcommand{\arraystretch}{1.3}
\caption{Success rates (SR) of published open-source methods and \ourmethod up to the completion of this work on \textsc{WebArena}, \textbf{Bold} represents the best result on the website while \underline{underline} means the second best results. The results originate from the corresponding papers except BrowserGym which we reproduce the GPT-4o version by ourselves. *: SteP \citep{sodhi2024stepstackedllmpolicies} uses human-designed detailed policies for each website, so it is not comparable with other autonomous methods without human involvement and we set it apart just for references.}
\begin{tabular}{lccccccc}
\toprule
Method & Shopping & CMS & Forum & Gitlab & Map & \textbf{Average} \\ \midrule
SteP* \citep{sodhi2024stepstackedllmpolicies} & 37.0 & 24.0 & 59.0 & 32.0 & 30.0 & 33.0 \\ \midrule
WebArena \citep{zhou2024webarenarealisticwebenvironment} & 24.0 & 11.0 & 7.9 & 10.2 & 21.1 & 15.0 \\
AutoEval \citep{pan2024autonomousevaluationrefinementdigital} & 25.5 & 18.1 & 25.4 & 28.6 & \textbf{31.9} & 20.2 \\  
BrowserGym \citep{drouin2024workarenacapablewebagents} & 26.6 & 28 & 22.8 & 21.4 & 18.4 & 24.3 \\ \hdashline
$\ourmethod_{offline}$ & \underline{29.2} & \underline{36.8} & 33.3 & \underline{36.7} & 29.5 & \underline{33.4} \\
$\ourmethod_{online}$ & \underline{29.2} & 36.3 & \underline{37.7} & 34.2 & 28.6 & 33.2 \\
$\ourmethod_{hybrid}$ & \textbf{32.8} & \textbf{41.2} & \textbf{41.2} & \textbf{37.2} & \underline{30.4} & \textbf{36.7} \\ \bottomrule
\end{tabular}
\label{tab:wa_main}
\end{table*}

\begin{table*}[t]
\centering
\renewcommand{\arraystretch}{1.3}
\caption{Success rates (SR) of published open-source methods and \ourmethod on \textsc{VisualWebArena}, \textbf{Bold} represents the best result in the domain. Results are from the corresponding papers except BrowserGym. We implement the agent with BrowserGym by ourselves. Due to cost consideration, we evaluate $\ourmethod_{online}$ only on \textsc{VisualWebArena}}.
\begin{tabular}{lcccc}
\toprule
Method      & Classifieds & Shopping & Forum & \textbf{Average} \\ \midrule
VisualWebArena \citep{koh2024visualwebarenaevaluatingmultimodalagents} & 18.4 & 20.0 & 17.1 & 18.9 \\
BrowserGym \citep{drouin2024workarenacapablewebagents} & 26.2 & 28.2 & 21.1 & 26.2 \\
Tree Search \citep{koh2024treesearchlanguagemodel} & 26.5 & 29.0 & 20.5 & 26.4 \\ \hdashline
$\ourmethod_{online}$ & \textbf{27.0} & \textbf{38.1} & \textbf{24.4} & \textbf{31.9} \\ \bottomrule
\end{tabular}
\label{tab:vwa_main}
\end{table*}

\subsubsection{WebArena}
\label{subsubsec:wa_details}
For \textsc{WebArena}, we evaluate all three settings of \ourmethod: offline, online and hybrid. In the offline setting, for training tasks, we first ask individuals unfamiliar with \textsc{WebArena} to navigate the website for three hours and design a few tasks they consider important. A filtering process is then applied to ensure that these tasks do not overlap with the test set. Finally, these tasks are annotated by humans to obtain gold trajectories for offline learning. The final tasks are listed in \ref{subsec:human_annotated_tasks_wa}. The data can also be annotated automatically by an LLM explorer. Due to space limitations, we discussed and compared these two offline data sources in \ref{subsec:off_on_analysis}.
We use GPT-4o-2024-0513 as the backbone language model with a temperature of 0.1. We use BrowserGym \citep{drouin2024workarenacapablewebagents} as the environment, which provides both text and visual observation for the agent and adds additional information for clickable and visible elements in the accessibility tree of the webpage. To fairly highlight the improvement of \ourmethod, we run GPT-4o w/ BrowserGym \citep{drouin2024workarenacapablewebagents} by ourselves as the baseline for comparison. 
\ourmethod is compatible with most off-the-shelf language model agents since it only needs the past trajectories. Here, we test it with a simple method by prompting GPT-4o directly and using ReAct \citep{yao2022react} as the output format as in BrowserGym \citep{drouin2024workarenacapablewebagents} and \textsc{WebArena} \citep{zhou2024webarenarealisticwebenvironment}. We also combine it with another performant method and observe significant improvements(\S\ref{subsec:synergy}). We set the retrieval parameter to $k_d = 5$ and $k_s = 5$, denoting the maximum number of dynamics/skills to retrieve and replay. 

\subsubsection{VisualWebArena}
For \textsc{VisualWebArena}, similar to \textsc{WebArena}, we still use BrowserGym \citep{drouin2024workarenacapablewebagents} as our environment. Since BrowserGym does not support visual evaluation, we implemented the environment by ourselves and built \ourmethod on top of that. We also run BrowserGym results as the baseline for comparison. Using the same setting as \citep{koh2024visualwebarenaevaluatingmultimodalagents}, we apply Set-of-Marks (SoM) \citep{yang2023setofmarkpromptingunleashesextraordinary} to the original screenshot of the webpage. This method marks each interactable element of the webpage with a highlighted bounding box and the corresponding unique element ID on the corner of the box to enable grounding. Besides the screenshot, the agent is also provided with text observation of the environment for better grounding, where the ID of each element is consistent with the one in the SoM-processed screenshot. 

\subsection{Results}

Our results on these two benchmarks are summarized in Table \ref{tab:wa_main} and Table \ref{tab:vwa_main}. On \textsc{WebArena} and \textsc{VisualWebArena}, while orthogonal to the other methods, \ourmethod achieves competitive performance and improves the baseline agent, GPT-4o w/ BrowserGym \citep{drouin2024workarenacapablewebagents}, relatively by 51.0\% ($\ourmethod_{hybrid}$) and 21.8\% ($\ourmethod_{online}$) respectively. It should be noted that SteP \citep{sodhi2024stepstackedllmpolicies} uses human-designed policies, i.e., step-by-step instructions for each website split, and can need much extra human effort when encountering new cases or on new websites. So we do not consider it when comparing \ourmethod with other methods. On \textsc{VisualWebArena}, \ourmethod achieves competitive performance and outperforms the tree search method \citep{koh2024treesearchlanguagemodel}, which is also built on GPT-4o, with much lower token costs. The result of the tree search is obtained through a search algorithm that uses 20 times sampling at each step, and a maximum of 5 steps, with extra costs of GPT-4o used as a value function. In our implementation, we use a maximum of only 30 steps for each task, similar to the setting in \citep{zhou2024webarenarealisticwebenvironment} and \citep{koh2024visualwebarenaevaluatingmultimodalagents}, thus using at least 3 times fewer tokens.

\begin{table*}[t]
\centering
\renewcommand{\arraystretch}{1.3}
\caption{Token efficiency and performance comparison between ReAct baseline and \ourmethod in three different settings. Since output tokens do not change much, we only count input tokens here.} 
\begin{tabular}{lcccc}
\toprule
Method & Success Rate & $\Delta$ Tokens & $\Delta$ Tokens (\%) & $\Delta$ SR (\%) \\ \midrule
ReAct (BrowserGym) & 24.3 & 0 & 0\% & 0\% \\
$\ourmethod_{offline}$ & 33.4 & 7885 & 5.8\% & 37.8\% \\
$\ourmethod_{online}$ & 33.2 & 16337 & 11.2\% & 36.7\% \\
$\ourmethod_{hybrid}$ & \textbf{36.7} & 20631 & 17.3\% & \textbf{52.5\%} \\ \bottomrule
\end{tabular}
\label{tab:cer_efficiency}
\end{table*}

\section{Analysis}
\label{sec:analysis}
In this section, 
we conduct extensive analysis to investigate and better understand \ourmethod's improvements through token efficiency, cross-template success rates and two interesting metrics for self-improvement systems: stability and plasticity. Finally, we validate its compatibility and synergy with other performant methods, proving its wide applicability.

\subsection{Token efficiency}
We have shown the effectiveness of our method in previous sections. Token efficiency, however, is also an important part to consider. We analyze the token cost of the ReAct baseline and CER under three different settings. Results are shown in Table \ref{tab:cer_efficiency} which shows that \ourmethod enables language agents to learn from past experiences efficiently (only minor additional costs, see table below) and effectively (significant improvement by 52.5\%). It shows that utilizing past experiences is essential for agents to do well and provide a simple but effective solution. Additionally, we also show in \ref{subsec:synergy} that CER can synergize well with other performant methods like Tree Search \citep{koh2024treesearchlanguagemodel} (we use the single-turn version of it for simplicity, which is sampling+reranking). This means that \textbf{with acceptable minor cost, CER can improve a base agent (e.g., ReAct or Tree Search) method easily and notably.}

\subsection{Investigating improvements of \ourmethod}
\label{subsec:improvement_analysis}

In this section, we try to understand where the improvements of \ourmethod come from and get some intuitions about how \ourmethod works.

Intuitively, the state space and action space for the current step are extremely large. However, for human users, the states that we often navigate to and the actions that we usually take are only a small subset of the whole space. Experiences distilled from some goal-oriented trajectories tend to contain some informative and effective states and actions that are often navigated to or used. With the highlighted promising states, actions, and decision-making patterns, the agent can issue a correct action much more easily. Of course, some of the experiences can be noisy and misleading. We show in section \ref{subsec:gtreward} that \ourmethod is still robust to the correctness of the trajectories. This intuition also aligns with the Recognition Primed Decision making Model proposed by \cite{Klein1999SourcesOfPower}, where humans tend to recognize promising actions when encountering complex environments.

We also conduct quantitative comparisons between \ourmethod and baseline method to investigate the improvements. The tasks in \textsc{WebArena} are designed based on templates, and at most, five tasks share the same template. If the agent just memorizes the pattern of the whole task, it will be able to solve some other tasks in the same template more easily, thus improving the overall performance. So, we use the cross-template average success rate, calculated by the number of templates solved (at least one task is solved) divided by the total number of templates. We run experiments on Forum tasks of \textsc{WebArena}. The results are shown in Table \ref{tab:wa_cross_template}. \ourmethod shows a significant improvement in cross-template success rates. This result validates that the improvement of \ourmethod does not come from memorizing the whole trajectory of a task. Instead, it distills more fine-grained experiences, which allows for the generalization of different types of tasks.
\begin{table}[t]
\centering
\renewcommand{\arraystretch}{1.3}
\caption{Cross-template success rates (ct-SR), stability and plasticity of \ourmethod and baseline on the Forum split of \textsc{WebArena}}
\begin{tabular}{lccc}
\toprule
 Method & ct-SR & Stability (\%) & Plasticity (\%) \\ \midrule
 Baseline & 44.7 & 100 & 100 \\ 
 \ourmethod & \textbf{60.0} & 93 & 141 \\ \bottomrule
\end{tabular}
\label{tab:wa_cross_template}
\end{table}


\subsection{Stability and Plasticity}
\label{subsec:stability}

A well-designed self-improvement system should demonstrate both stability (preservation of old knowledge) and plasticity (acquisition of new knowledge) \citep{Grossberg1982CognitiveCode, rolnick2019experiencereplaycontinuallearning}. Since knowledge is hard to measure in our case, we measure the acquisition of new knowledge through problem-solving ability, i.e., success rates, in a specific environment. Therefore, we similarly measure the stability and plasticity of \ourmethod in cross-template success rate (ct-SR) since the success in new types of task demonstrates new problem-solving ability. Specifically, stability is measured through the percentage of tasks from the baseline that CER is able to solve, which reflects how well CER maintains the original capability of the baseline. Plasticity is measured by the
improvement of CER on new cases, measuring how many additional tasks CER can solve compared
to the baseline. Since \ourmethod can be understood as a self-improvement system built on the baseline method agent. We set the stability and plasticity of the baseline to 100\% and used the ct-SR to calculate the stability and plasticity of \ourmethod. The results are also shown in Table \ref{tab:wa_cross_template}. With most of the original abilities retained, \ourmethod demonstrates 41\% new problem types solved, proving the validity of \ourmethod as a self-improvement framework. This also indicates the potential of the compatibility with other performant methods, which we discuss in detail in section \ref{subsec:synergy}.

\subsection{Validity on open-source models}
\begin{table}[t]
\centering
\renewcommand{\arraystretch}{1.3}
\caption{Comparison of success rates (SR) of ReAct and ReAct w/ \ourmethod with Llama-3.1-70B on the Gitlab split of \textsc{WebArena}}
\begin{tabular}{lc}
\toprule
 Method & SR\\ \midrule
 Baseline (ReAct) & 17.3 \\
$\ourmethod_{hybrid}$ & 22.0 \\ \bottomrule
\end{tabular}
\label{tab:llama_results}
\end{table}
We also evaluate \ourmethod on a weaker open-source model: Llama-3.1-70B \cite{grattafiori2024llama3herdmodels}. We evaluate the ReAct baseline and \ourmethod on Gitlab split (the largest split on \textsc{WebArena}). As shown in Fig. \ref{tab:llama_results}, \ourmethod still improves the baseline relatively by 26.53\%, proving its validity on weaker open-source models. However, the improvement is smaller than it is with strong models like GPT-4o. We also analyzed the results and found that weaker models like llama3.1 do worse compared with GPT-4o at formatting their output when solving challenging tasks like \textsc{WebArena} with a larger action space. This also influence their robustness when distilling some multi-step, useful and well-formatted skills, which explains why the improvement is relatively smaller than it does on strong models like GPT-4o.

\subsection{Synergy with performant methods}
\label{subsec:synergy}

We also analyze the compatibility of \ourmethod with other performant methods. Tree search \citep{koh2024treesearchlanguagemodel} is a computing-intensive method with more explorations and backtracking to search for better action to take. However, due to the high costs and the long time it takes, we chose another comparable method of it: trajectory sampling and reranking. We sampled 3 times for each task with max steps of 20 and prompted a language model with the trajectories to give a score and select the trajectory with the highest score as the final one. The procedure of applying \ourmethod to such method is similar to what we do with baseline agent. We conduct experiments in an online setting on the Forums split of \textsc{WebArena}. The results are in Table \ref{tab:samping_ablation}. \ourmethod with sampling method improves \ourmethod w/ ReAct performance by a relative success rate increase of 39.5\%.
This is because, firstly, such performant methods generally have better precision from start to end, so the experiences distilled from them are of higher quality. Additionally, high-quality experiences make performant methods more robust through the learned experiences and provide environment-specific knowledge to help better decision-making. 
\begin{table}[t]
\centering
\renewcommand{\arraystretch}{1.3}
\caption{Comparison of success rates (SR) of \ourmethod and \ourmethod w/ trajectory sampling and reranking on the Forum split of \textsc{WebArena}}
\begin{tabular}{lc}
\toprule
 Method & SR\\ \midrule
 \ourmethod & 37.7 \\
 Sampling & 43.1 \\ \hdashline
 \ourmethod w/ sampling & \textbf{52.6} \\ \bottomrule
\end{tabular}
\label{tab:samping_ablation}
\end{table}

\subsection{Access to ground truth rewards}
\label{subsec:gtreward}
\begin{table}[h]
\centering
\renewcommand{\arraystretch}{1.3}
\caption{Success rates (SR) of $\ourmethod$ and $\ourmethod_{success}$ on  \textsc{WebArena}. $\ourmethod_{success}$ uses ground truth evaluators from the environment to filter out and learn from successful experiences only. Both method takes text observation for comparison}.
\begin{tabular}{@{}lllllll@{}}
\toprule
Method     & \textbf{Average} \\ \midrule
\ourmethod & 31.4 \\
$\ourmethod_{success}$ & \textbf{33.5} \\ \bottomrule
\end{tabular}
\label{tab:wa_reward_ablation}
\end{table}
Currently, \ourmethod distills experiences from both successful and failed cases. To investigate whether the distillation from failed cases is a bottleneck of \ourmethod, we run \ourmethod for only successful trajectories evaluated by ground truth evaluators. We conduct the experiments on the full set of \textsc{WebArena}. The results are summarized in Table \ref{tab:wa_reward_ablation}.
The results show that \ourmethod performs better with the access to ground truth reward. The possible reason could be that the successful trajectories have higher quality and are more informative, while the failed trajectories provide some misleading action sequences that will negatively influence the agent's decision-making.

Nevertheless, the acceptable gap and significant improvements over the baseline agent show the robustness of \ourmethod given noisy trajectories. This gives credit to the implicit reasoning ability and the flexible natural language representation of experiences. Although provided with a few noisy experiences, the agent can still filter out useful trajectories and issue reasonable action mostly.

\subsection{Division of Dynamics and skills}
\label{subsec:module_ablation}

We conduct an ablation experiment to investigate the necessity of the division of dynamics and skills. The experiment is run on the Forums split of WebArena. The results in Table \ref{tab:wa_module_ablation} show that both dynamics and skills are important for \ourmethod. Environment dynamics make the agent aware of the content of many pages and also provide the URL to navigate to. Skills inspire the agent and also provide promising actions to be taken in the current step. They provide heuristics in terms of states and actions correspondingly and synergize with each other.
\begin{table}
    \centering
\renewcommand{\arraystretch}{1.3}
\caption{Success rates (SR) of \ourmethod on the Forum split of WebArena with different ablation settings to the experiences.}
\resizebox{0.28\textwidth}{!}{
\begin{tabular}{lc}
\toprule
 Method & SR \\ \midrule
 \ourmethod & \textbf{37.7} \\ 
 \ourmethod - skills & 33.3 \\ 
 \ourmethod - dynamics & 35.1 \\ \bottomrule
\end{tabular}
}
\label{tab:wa_module_ablation}
\end{table}
\section{Conclusions}
In this paper, we proposed a training-free framework for efficient and effective self-improvement of language agents in complex web environments. Our framework enables language agents to learn from past experiences and replay during inference time for better decision-making. We also conduct extensive analysis to investigate its improvements and validate its effectiveness as a self-improvement system through stability and plasticity. We believe that learning from past experiences is crucial for building a helpful computer agent that can adapt to different environments and evolve autonomously.

\section{Limitations}
Despite the substantial progress achieved with \ourmethod, there are several limitations that will influence its applicability and could be addressed in future work. First, table \ref{tab:wa_main} shows that although \ourmethod performs even better in offline + online settings, it requires the trajectories to be goal-oriented to distill high-quality experiences. The performance is limited if trajectories from random explorations are provided. The more fine-grained utilization of low-quality trajectories could be explored in the future. Secondly, the environment dynamics help much in web environments, as shown in section \ref{subsec:module_ablation}, partially because the agent can directly navigate to a specific page with its URL. It would be interesting to investigate how we can utilize the environment dynamics in other agent tasks, such as real-world navigation \citep{shridhar2021alfworld} in future work. Also, CER needs to learn trajectory data before starting the task. In online setting, the framework may not be able to use trajectory learning in the initial decision. In offline setting, the human-annotated trajectory will be used for CER learning, but the trajectory data prepared in advance depends on the annotation of the human annotator. Hybrid is a good improvement solution, but it still can hardly avoid dependence on the human annotator. How to generate such trajectories in a more efficient way would be an interesting topic to dive into in the future.

\section{Ethical and Broader Impacts}
\subsection{Real World Impacts}
Advancing the capabilities
of autonomous agents comes with many broader
considerations and ethical implications. Strong autonomous agents have the potential to improve the
accessibility of computer-based tasks, potentially
aiding individuals with disabilities or those lacking technical skills. More broadly, agents have
the potential to automate large portions of routine
computer work. While the capabilities of existing
autonomous agents are insufficient for even simple tasks (as shown in this paper), these impacts
highlight the need to ensure that the broader economic and social implications on employment are
carefully considered if/when autonomous agents
are deployed in real world applications.
\subsection{Bias and Safety}
When developing autonomous
agents, it is also imperative to ensure that these
agents do not inadvertently exclude or disadvantage
any group. Further analysis is essential to ensure
that deployed agents do not exhibit unintended biases. Agents also have the potential to cause more
harm (than regular LLMs) in real world applications if careful safeguards are not in place. Further
research is necessary to understand and mitigate
possible harmful behaviors.
\subsection{Intended Uses}
Our method is based on \textsc{WebArena} and \textsc{VisualWebArena} which are research
benchmarks to measure and evaluate the progress
of multimodal agents. The models and methods we presented in
this paper are research prototypes, and not intended
for deployment in practical applications in their
current state (especially in high risk domains).

\appendix

\section{Appendix}
\subsection{\ourmethod prompts}
\label{subsec:cer_prompts}
Here we provide detailed prompts for each module in \ourmethod. Figure \ref{fig:dynamics_dis_system_message} and \ref{fig:skill_dis_system_message} show the system prompts for distillation modules, while Figure \ref{fig:retrieval_dis_system_message} and \ref{fig:retrieval_skills_system_message} are system prompts for retrieval modules.
\begin{figure*}[th]
\noindent\fbox{\parbox{\textwidth}{%
\small
You will be given the state-action trajectory of a user interacting with a webpage and the overall goal of the trajectory.
You need to summarize the useful pages and pair it up with the corresponding URLs.

Output format: \\
$<$URL$>$ \\
the URL of page 1 \\
$<$/URL$>$ \\
$<$think$>$ \\
think step by step like in the examples and summarize the page \\
$<$/think$>$ \\
$<$page-summary$>$ \\
the brief summary of page 1, following the format: \\
Name: {{name}} page \\
Description: {{descriptions}} \\
Usages: {{usages}} \\
$<$/page-summary$>$ \\
$<$URL$>$ \\
the URL of page 2 \\
$<$/URL$>$ \\
$<$think$>$ \\
think step by step like in the examples and summarize the page \\
$<$/think$>$ \\
$<$page-summary$>$ \\
the brief summary of page 2, following the format: \\
Name: {{name}} page \\
Description: {{descriptions}} \\
Usages: {{usages}} \\
$<$/page-summary$>$ \\
... \\

\# Examples \\
\#\# Example 1 \\
Overall goal of the trajectory: Go to r/books forum. \\ 
Current website: Reddit \\
Existing summarized pages: \\
Page 1: Profile page \\
Description: it shows the user's profile information. \\
Usages: view or modify user's profile information. \\
URL: https://www.example.com/profile \\
Human user trajectory: [neglected here] \\

\#\# Output: \\
$<$URL$>$ \\
https://www.example.com/forums \\
$<$/URL$>$ \\
$<$think$>$ \\
From the content of the page, it shows a list of forums, I can summarize it to Forum page. The website is Reddit so this page can be used to navigate to different forums. \\
$<$/think$>$ \\
$<$page-summary$>$ \\
Name: Forums page; Description: it shows a list of different forums.; Possible usages: navigate to different forums. \\
$<$/page-summary$>$ \\

IMPORTANT NOTES you should absolutely follow: \\
1. DO NOT include any other words except url, think and page summary as the format stated above. \\
2. Follow the example to think and summarize the page. \\
3. You should only summarize once for each unique URL. \\
4. Check existing pages before generating, do not summarize pages that have already been summarized, instead, use "Summarized before" in the steps. \\
5. Focus on the main content of the page and may ignore the modifications made by the user when generating the summary. \\
}
}
\caption{System message for dynamics distillation module in \ourmethod}  \label{fig:dynamics_dis_system_message}
\end{figure*}
\begin{figure*}[th]
\noindent\fbox{\parbox{\textwidth}{%
\small
You will be given the state-action trajectory of a user interacting with a webpage and the overall goal of the trajectory.\\
You need to summarize skills from the trajectory.\\
Skills are a subset of actions that the user takes to achieve a sub-goal.\\
You should break the overall goal into sub-goals and summarize each sub-goal as a skill.\\
Represent the non-fixed elements (input text, button strings) and non-fixed words (e.g. a specific forum name / user name; an option) with descriptive variable names as shown in the example.\\
Output format:\\
$<$think$>$\\
think step by step\\
$<$/think$>$\\
$<$skill$>$\\
skill1 name here.\\
$<$/skill$>$\\
$<$steps$>$\\
The steps of the skill1 here.\\
$<$/steps$>$\\
$<$think$>$\\
think step by step\\
$<$/think$>$\\
$<$skill$>$\\
skill2 name here.\\
$<$/skill$>$\\
$<$steps$>$\\
The steps of the skill2 here.\\
$<$/steps$>$\\
...\\
\# Examples\\
\#\# Example 1\\
Overall goal: I want to get the cheapest product in the Cabinets, Racks \& Shelves category\\
Current website: {{current website}}\\
Existing skills:\\
Skill 1: Sort products by {{sort criterion}}\\
1. To sort the products by {{sort criterion}}, I need to click on the "Sort by" dropdown menu.\\
```click({{sort by id}})```\\
2. To sort the products by {{sort criterion}}, I need to select the {{sort criterion}} option from the "Sort by" dropdown menu.\\
```click({{sort criterion id}})```\\
Human user trajectory: [neglected here for length]\\
\#\#Output: [neglected here for length]\\
IMPORTANT NOTES you should absolutely follow: \\
1. DO NOT include any other words except skills and steps as the format stated above.\\
2. Check existing skills before generating; do not summarize skills that have already been summarized; instead, use "Summarized before" in the steps.\\
3. You should break the overall goal into sub-goals and summarize each sub-goal as a skill.}}
\caption{System message for skills distillation module in \ourmethod}  \label{fig:skill_dis_system_message}
\end{figure*}
\begin{figure*}[th]
\noindent\fbox{\parbox{\textwidth}{%
\small
You will be given a goal of a task to be executed on a website and a list of urls and the corresponding page summary to choose from.\\
You need to select the pages that most possibly need to be visited to achieve the goal.\\
You should break the task down into a few steps so that you can select the pages that can help most in each step.\\
IMPORTANT: You should select not more than {max skills} pages!\\
Output format:\\
$<$think$>$\\
think step by step. Break the task down into a few steps and then select pages\\
$<$/think$>$\\
$<$selected-pages$>$\\
id: {{the id number (the number at the beginning) of page 1}}; name: page 1 name\\
id: {{the id number (the number at the beginning) of page 2}}; name: page 2 name\\
...\\
$<$/selected-pages$>$\\
\\
\# Examples\\
\#\# Example 1\\
Task goal: Upvote the hottest post in r/books\\
Current website: {website descriptions}\\
Shortcuts to choose from:\\
id: 1; name: Forums page; description: It shows a list of different forums; possible usages: navigate to different forums; url: https://www.example.com/forums\\
id: 2; name: Profile page; description: It shows the information of current user; possible usages: Check or modify the information of the current user; url: https://www.example.com/profile\\
id: 3; name: Submission page; description: It provides a few text boxes to fill in to submit a new post; possible usages: Submit new posts; url: https://www.example.com/submission\\
id: 4: name: Subscribed forums page; description: It provides a list of subscribed forums; possible usages: check or navigate to subscribed forums; url: https://www.example.com/subscribed\\
\#\# Output 1:\\
$<$think$>$\\
The goal is to upvote the hottest post in r/books. The user needs to navigate to the r/books page first or go to forums to find the r/books page. Then the user needs to find the hottest post in the r/books page. So the useful pages from the shortcuts are Forums page\\
$<$/think$>$\\
$<$selected-pages$>$\\
id: 1; name: Forums page\\
$<$/selected-pages$>$}}
\caption{System message for dynamics retrieval module in \ourmethod}  \label{fig:retrieval_dis_system_message}
\end{figure*}
\begin{figure*}[th]
\noindent\fbox{\parbox{\textwidth}{%
\small
You will be given a goal of a task to be executed on a website and a list of skills to choose from.\\
You need to select the skills that can help most in achieving the goal.\\

You should break the task down into a few steps so that you can select the skills that can help most in each step.\\
IMPORTANT: You should select not more than {max skills} skills!\\
Output format:\\
$<$think$>$\\
think step by step. Break the task down into a few steps and then select skills\\
$<$/think$>$\\
$<$selected-skills$>$\\
id: {{the id number (the number at the beginning) of skill 1}}; name: skill 1 name\\
id: {{the id number (the number at the beginning) of skill 2}}; name: skill 2 name\\
...\\
$<$/selected-skills$>$\\
\# Examples\\
\#\# Example 1\\
\\
Task goal: Upvote the hottest post in r/books\\
Current website: {website descriptions}\\
Skills to choose from:\\
Skill 1: Navigate to forums\\
1. Click on the "Forums" menu item.\\
```click({{forums id}})```\\
2. Click on the specific forum name.\\
```click({{forum name id}})```\\
Skill 2: Submit a new post\\
1. Type the post title in the title text box.\\
```type({{title text box id}}, "Post Title")```\\
2. Type the post content in the content text box.\\
```type({{content text box id}}, "Post Content")```\\
3. Click on the "Submit" button.`\\
``click({{submit button id}})```\\
Skill 3: Sort posts by {{sort criterion}}\\
1. Click on the "Sort by" dropdown menu.\\
```click({{sort by dropdown id}})```\\
2. Select the {{sort criterion}} option from the "Sort by" dropdown menu.\\
```click({{sort criterion id}})```\\
\\
Output:\\
$<$think$>$\\
The goal is to upvote the hottest post in r/books. The user needs to navigate to the r/books page first or go to forums to find the r/books page. Then the user needs to find the hottest post in the r/books page. So the useful skills from the shortcuts are Navigate to forums, Sort posts by hotness\\
$<$/think$>$\\
$<$selected-skills$>$\\
id: 1; name: Navigate to forums\\
id: 3; name: Sort posts by hotness\\
$<$/selected-skills$>$\\
\\
Notes:\\
1. Some skills might not be consistent with the current task but it is still useful to refer to, e.g. write a post to express happiness is useful in a task to write a post to express sadness.}}
\caption{System message for skills retrieval module in \ourmethod}  \label{fig:retrieval_skills_system_message}
\end{figure*}
\subsection{Experience examples}
\label{subsec:experience_examples}
We provide experiences examples for \textsc{WebArena} and \textsc{VisualWebArena} here. These natural language experiences are programmatically transformed from the structured experiences stored in the experience buffer and will be added to the system prompt of the model when solving tasks. The examples are shown in Figure \ref{fig:wa_experience_example} and \ref{fig:vwa_experience_example}.
\begin{figure*}[h]
\noindent\fbox{\parbox{\textwidth}{%
\small
\# Environment dynamics (common pages to navigate to from previous experience): \\
\#\# Dynamics 1: Forums page \\
\#\#\# Forums page contents: it shows a list of different forums. \\
\#\#\# Potential usages: navigate to different forums. \\
\#\#\# URL: http://localhost:9999/ \\
... \\
\# Skills (common workflows summarized from previous experience): \\
\#\# Skill 1: Edit profile biography \\
\#\#\# Steps: \\
1. Navigate to the Edit Biography page. \\
```goto('http://localhost:9999/user/{username}/edit biography')``` \\
2. Fill in the biography text area with the new biography content. \\
```fill({biography text area id}, '{new biography content}')``` \\
3. Click on the save button to update the biography. \\
```click({save button id})``` \\
...
}}
\caption{Experience snippet example on \textsc{WebArena}}  \label{fig:wa_experience_example}
\end{figure*}
\begin{figure*}[h]
\noindent\fbox{\parbox{\textwidth}{%
\small
\# Environment dynamics (common pages to navigate to from previous experience): \\
\#\# Dynamics 1: Video Games category page \\
\#\#\# Video Games category page contents: This page lists products that are within the "Video Games" category, including accessories, consoles, and other related products \\
\#\#\# Potential usages: Browse and purchase video game-related products. \\
\#\#\# URL:http://localhost:7770/video-games.html \\
... \\
\# Skills (common workflows summarized from previous experience): \\
\#\# Skill 1: Sort search results by \{sort criterion\} \\
\#\#\# Steps: \\
\#\#\# 1. Open the sort dropdown menu by clicking on the sort dropdown button. \\
```click(\{sort dropdown button id\})``` \\
2. Select the "\{sort criterion\}" option from the sort dropdown menu. \\
```select\_option(\{sort dropdown button id\}, \{sort criterion\})``` \\
...}}
\caption{Experience snippet example on \textsc{VisualWebArena}}  \label{fig:vwa_experience_example}
\end{figure*}
\subsection{Self-guided exploration prompt}
\label{subsec:exploration_prompt}
As mentioned in section \ref{subsec:off_on_analysis}, we prompt a language model, i.e., GPT-4o, here to explore diverse actions in the environment and collect the corresponding trajectories. We limit the max steps to 30 and sample 10 trajectories. The instruction part of the prompt is shown in Figure \ref{fig:rd_explore_instruction_prompt}.
\begin{figure*}[h]
\noindent\fbox{\parbox{\textwidth}{%
\small
\# Instructions:
Your objective is to discover diverse and interesting tasks (that a human might give to an agent) by interacting
with the webpage through these actions. You’ve executed the following actions, and observed the following webpage
states 
[observations and other information are neglected here]
}}
\caption{The instruction part of prompt for random explore agent}  \label{fig:rd_explore_instruction_prompt}
\end{figure*}
\subsection{Human annotated tasks for \textsc{WebArena}}
\label{subsec:human_annotated_tasks_wa}
As stated in section \ref{subsubsec:wa_details}, we asked individuals unfamiliar with \textsc{WebArena} to design a few common tasks for each website in \textsc{WebArena} and asked them to annotate the oracle trajectory for each task. The number of training tasks varies across websites, ranging from five to ten, and accounts for approximately 5\% of the test set. The full list of task instructions for each website is as follows:
\subsubsection{Shopping Admin}
\begin{itemize}
    \item \textit{How many customers expressed they were impressed by the product}
    \item \textit{Tell me the best-selling products each year from 2022 to now}
    \item \textit{Change the street address of the order with ID 300 to 790 Harvard Square}
    \item \textit{Tell me the number of canceled orders in total. Tell me the users who have canceled orders ever}
    \item \textit{Tell me the customer who made the most orders during Jan 2022}
    \item \textit{Tell me the month when there are the most orders in 2022}
    \item \textit{Modify the product Atlas fitness tank L blue’s stock quantity to 80}
    \item \textit{Tell me Sophia Kim's email address}
    \item \textit{Tell me the number of pending reviews in total}
    \item \textit{How does Merrie feel about the Antonia tank?}
\end{itemize}
\subsubsection{Shopping}
\begin{itemize}
    \item \textit{How many complete orders are there in August 2022}
    \item \textit{How many items have I bought in total in Sep 2022}
    \item \textit{What is the size of pancakes I bought on April 5, 2022}
    \item \textit{What is the price of the most expensive toothpaste}
    \item \textit{What is the price of the cheapest coffee}
    \item \textit{I want to buy some energy drinks now, help me find the blueberry flavor ones and a pack of 24.}
    \item \textit{Are there any out-of-stock orders there?} 
    \item \textit{Go to my first completed order in history}
    \item \textit{Go to the bread \& bakery page.}
    \item \textit{Find the V8 energy drink and tell me the number of reviews of it about “good taste”}
\end{itemize}
\subsubsection{Gitlab}
\begin{itemize}
    \item \textit{Tell me the title of an issue with a label about bugs and the top priority in the repo “administrate”}
    \item \textit{Directly type everything during filtering}
    \item \textit{Who made the latest commit in repo “ArduinoJson”}
    \item \textit{Tell me how many merge requests I need to review}
    \item \textit{Maybe need to tell the agent I am ByteBlaze?}
    \item \textit{Tell me what is the license of repo “chinese-colors”}
    \item \textit{What is the number of members in repo “css-selector”}
    \item \textit{Create a private project from iOS template with project name “abc”}
    \item \textit{Create an issue with “Setup problems” as the title in repo “ArduinoJson”}
    \item \textit{Who has the most contribution to project “xssor2”}
    \item \textit{Assign “Support linking to an accessibility statement” issue inside my a11y-webring.club repo to Rohan}
    \item \textit{What is the highest number of stars a repository has ever received on GitLab?}
\end{itemize}

\subsubsection{Map}
\begin{itemize}
    \item \textit{What is the walking time from Friend Center to Frist Campus Center at Princeton University?}
    \item \textit{Tell me whether I should walk or drive to Frist Campus Center to get there faster from Friend Center at Princeton University?}
    \item \textit{How long does it take to go from Princeton University to where Independence Hall lies by driving?}
    \item \textit{Find details about the Independence Hall.}
    \item \textit{Show me the list of hospitals close to Princeton University.}
\end{itemize}

\subsubsection{Forum}
\begin{itemize}
    \item \textit{Find the title of the most commented post in forum 'history'.}
    \item \textit{Find the title of the most controversial post of all time in forum 'Paterson'}
    \item \textit{Go to the user's profile who made the comment "What are you doing in a deep learning sub?"}
    \item \textit{Upvote the comment of the current user which replies to 'Maybe there's just something wrong with me.' by HedleyLamarrrr}
    \item \textit{Reply to the first comment of the post titled 'It's the only logical explanation' with 'lol'}
\end{itemize}
\subsection{Offline data sources}
\label{subsec:off_on_analysis}
In section \ref{subsubsec:wa_details}, $\ourmethod$ uses human annotated data for offline learning. We conduct offline and offline + online experiments on the Forum tasks split of WebArena with two sources of offline training trajectories: from human demonstrations or from self-guided explorations to investigate their effects.

For the human annotation data, We use the same tasks as in section \ref{subsubsec:wa_details}. The details can be found in section \ref{subsec:human_annotated_tasks_wa}. For the random exploration data, we prompt a language model to propose diverse actions at each step and collect the final exploration trajectories. The details of prompts and tasks can be found in \ref{subsec:exploration_prompt} and \ref{subsec:human_annotated_tasks_wa}. 
\begin{table}[t]
\centering
\renewcommand{\arraystretch}{1.3}
\caption{Success rates (SR) of different settings with different offline data source on the Forum tasks split of the WebArena.}
\begin{tabular}{clc}
\toprule
 & Offline data source & SR\\ \midrule
 Baseline & - & 22.8 \\ \hdashline
\multirow{2}{*}{Offline} & human annotations & 33.3 \\
& self-guided explorations & 31.6 \\ \hdashline
Online & - & 37.7 \\ \hdashline
\multirow{2}{*}{Hybrid} & human annotations & \textbf{41.2} \\
& self-guided explorations & 35.1 \\ \bottomrule
\end{tabular}
\label{tab:wa_offline_ablation}
\end{table}
The overall results are shown in Table \ref{tab:wa_offline_ablation}. Both training sources improve the performance over the baseline in the offline setting. Notably, with five human annotations as the training set, the performance of offline + online learning surpasses the original online learning. To understand how offline and online learning synergize with each other, we take task 31 as an example: the agent is asked to get the count of comments that have received more downvotes than upvotes for the user who made the latest post on the photoshopbattles forum. In online settings, the agent does not know how to sort the posts on the Forum website either because this task is at the beginning, and it has not learned many experiences yet. However, from the training set, \ourmethod distilled the page summary of the forums page where all forums are displayed and also the skill of sorting the list of posts. Aware of the existence of the forums page, the agent knows to click the "forums" button to navigate to the list of forums. After that, it is inspired by the skill and sorts the posts correctly. It finishes the task successfully with dynamics and skills distilled from the training set.

Although in the offline setting, both training sources help the agent outperform the baseline, offline + online settings with self-guided explorations perform even worse than the online-only setting. We analyzed the results and found that the trajectories collected through such explorations are highly unstructured and noisy. The exploration agent can jump from one action to another unrelated one. So, the distillation module can hardly distill useful patterns from it. So, the distilled experiences may even mislead the agent sometimes.

In real-world scenarios, high-quality human-annotated training data is hard to collect, so online learning is still important and meaningful in most cases. It would be interesting to explore the potential of \ourmethod with more high-quality human-labelled trajectories. The negative impact of the training set derived from explorations also indicates that goal-oriented trajectory matters for \ourmethod because it has more structured, continuous, and relatively meaningful action sequences rather than unordered small pieces.

\end{document}